# Toward Optimal Adenovirus Detection Using YOLO26

**Olivier Rukundo**

Department of Electronic and Computer Engineering; University of Limerick; Limerick, Ireland
e-mail: olivier.rukundo@ul.ie

**Abstract:** This study systematically benchmarks different data augmentation setups across the baseline YOLO26 model size variants to determine the most effective setup for adenovirus detection in TEM images. The benchmarked setups include NAS, GAS, GMAS, and DAS, all evaluated under identical training conditions. A modified YOLO26 model leveraging P2, expanded STAL, increased *topk*, and MuSGD was also tested across the same benchmarked setups. The adenovirus dataset, selected from the published TEM virus dataset, was re-annotated by leveraging adenovirus particle positions to generate YOLO-compatible bounding box annotations. The modifications produced their largest performance gains under GAS and GMAS, with modified YOLO26x trained using GAS achieving a *mAP@50* of 0.80, *Precision* of 0.81 and a *Recall* of 0.80

## 1. Introduction

Transmission electron microscopy (TEM) remains one of the most reliable techniques for directly visualizing adenoviruses, as large-scale production of viral vectors for gene therapy requires tools to characterize such virus particles [1-2]. With TEM enabling nanometer-scale resolution, it has become possible to understand what occurs with viral particles when parameters or process operations change or when formulations are modified[3-4]. However, detecting accurately adenoviruses in TEM images remains challenging due to the presence of broken adenoviruses, debris, and various kinds of staining artefacts [4]. Deep learning methods have demonstrated strong potential for automated adenovirus detection [3–4]. However, despite the success of YOLO-based detectors in numerous object detection tasks [5], their application to adenovirus detection has not been previously investigated. Recently, YOLO26 was introduced as the latest member of the YOLO family [6], providing an opportunity to investigate its applicability to detect adenovirus in TEM images from TEM virus dataset [7]. Despite advances in object detection architectures [8-12], the influence or impact of data augmentation setups on adenovirus detection remains largely unexplored. For the purposes of this study, a data augmentation setup is defined as a predefined set of data augmentation transformations and their corresponding parameter settings applicable to the training data. Because a data augmentation setup can substantially affect object detection model generalization, identifying the most effective augmentation setup is essential for reliable adenovirus detection in TEM images, especially, due to the simultaneous presence of intact adenoviruses, broken adenoviruses, different types of debris, and various kinds of staining artefacts in such TEM images [4]. Therefore, this work systematically benchmarks multiple data augmentation setups across baseline YOLO26 model size variants to determine the most effective augmentation setup for adenovirus detection. The benchmarked data augmentation setups are: No Augmentation Setup (NAS), Geometric Augmentation Setup (GAS), Geometric + Mosaic Augmentation Setup (GMAS), and Default Augmentation Setup (DAS). The selected YOLO26 model size variants are *nano* (n), small (s), medium (m), large (l) and extra-large (x) [6]. Furthermore, because general-purpose baseline object detectors are not specifically optimized for every application, a modified YOLO26 detector integrating the P2 feature level, expanded Small Target Label Assignment (STAL), increased *top-k* positive-sample assignment, and MuSGD was developed to improve the detection of small (adenovirus) particles and better exploit the evaluated data augmentation setups. The contributions of this work are threefold. First, four data augmentation setups (DAS, NAS, GAS, and GMAS) were systematically benchmarked across all five YOLO26 model size variants under identical experimental conditions. Second, a modified YOLO26 detector integrating the P2 feature level, an expanded STAL strategy, increased *top-k* positive-sample assignment, and the MuSGD optimizer was developed to improve the detection of small adenovirus particles and better exploit the evaluated data augmentation setups. Third, the experimental evaluation identified the most effective data augmentation setup and demonstrated the effectiveness of the proposed baseline detector modifications.

## 2. Materials and Methods

### *2.1. Data Augmentation Setups*

In Deep learning, the data augmentation relies on geometric transformations (such as rotation, scaling, translation, flipping), non-geometric transformations (such as contrast adjustment, blurring, noise addition) and composite transformations or multi-image composition methods (such as Mosaic) [13-15]. The most popular are geometric transformations which rely heavily on interpolation algorithms to estimate pixel values at transformed coordinates to preserve image continuity after spatial transformation. Early studies investigated image interpolation from a general perspective, establishing a broad understanding of grayscale interpolation methods [16].

Table 1: Data augmentation setups evaluated.

| Transformation | DAS | NAS | GAS | GMAS |
|---|---|---|---|---|
| Rotation | Default | 0.0 | 180.0 | 180.0 |
| Translation | Default | 0.0 | 0.10 | 0.10 |
| Scaling | Default | 0.0 | 0.20 | 0.20 |
| Shear | Default | 0.0 | 0.0 | 0.0 |
| Perspective | Default | 0.0 | 0.0 | 0.0 |
| Horizontal flip | Default | 0.0 | 0.5 | 0.5 |
| Vertifical flip | Default | 0.0 | 0.5 | 0.5 |
| Mosaic | Default | 0.0 | 0.0 | 1.0 |
| Mixup | Default | 0.0 | 0.0 | 0.0 |
| Copy-Paste | Default | 0.0 | 0.0 | 0.0 |
| HSV-Hue | Default | 0.0 | 0.0 | 0.0 |
| HSV-Saturation | Default | 0.0 | 0.0 | 0.0 |
| HSV-Value | Default | 0.0 | 0.0 | 0.0 |
| Close Mosaic | Default | 0.0 | 0.0 | 10 |

Subsequent studies examined the influence of rounding functions on interpolation accuracy, demonstrating that interpolation performance can be significantly affected by the choice of rounding strategy [17–19]. A new conceptual framework for classifying interpolation algorithms into extra-pixel and non-extra-pixel methods was later introduced [20]. Building upon these findings, subsequent research focused on developing improved interpolation algorithms through adaptive optimization, novel weighting schemes, and extrapolation techniques [21–26]. These developments were ultimately extended to practical medical imaging applications, where optimized interpolation was employed to improve echocardiographic scan conversion [19,27]. In this context, the interpolation algorithms play a fundamental role in data augmentation pipelines used in deep learning methods.

As of today, the nearest-neighbour, bilinear, and bicubic interpolations are still among the most commonly used interpolation algorithms for geometrically transforming images during data augmentation [13]. Geometric data augmentation applies affine transformations to both the input image and the corresponding bounding-box or pixel level annotations. The general form of affine transform is shown in Eq. (1)

$$[x \quad y \quad 1] = [v \quad w \quad 1]\, T \tag{1}$$

where

$$T = \begin{bmatrix} t_{11} & t_{12} & 0 \\ t_{21} & t_{22} & 0 \\ t_{31} & t_{32} & 1 \end{bmatrix} \tag{2}$$

This transformation can scale, rotate, translate, or sheer a set of coordinate points, depending on the value chosen for the elements of matrix $T$ [34]. More details on the Eq.(2)'s matrix values used to implement these transformations are provided in [34]. Since the transformed coordinates generally do not coincide with integer pixel locations, the transformed pixel intensity is estimated using an interpolation algorithm, especially the bilinear interpolation, whose equation is given in Eq. (3)

$$\begin{aligned} I(x', y') = {} & I(x_0, y_0)(1-\alpha)(1-\beta) \\ & + I(x_1, y_0)\alpha(1-\beta) \\ & + I(x_0, y_1)(1-\alpha)\beta \\ & + I(x_1, y_1)\alpha\beta, \end{aligned} \tag{3}$$

where $\alpha$ and $\beta$ denote the fractional horizontal and vertical distances, respectively while $I(x_i, y_j)$ denotes the intensity of a neighbouring pixel [17], [23]. In the implementation adopted in this study, the above bilinear interpolation function was realized through the standard Ultralytics data augmentation pipeline. According to the Ultralytics implementation, image resizing is performed using OpenCV linear interpolation (cv2.INTER_LINEAR). Geometric transformations are implemented using OpenCV affine and perspective transformation functions (cv2.warpAffine and cv2.warpPerspective), which use linear interpolation (cv2.INTER_LINEAR) by default. However, the choice of interpolation algorithm remains dependent on the image type and/or the intended application [13]. For example, for semantic segmentation, nearest-neighbour interpolation (cv2.INTER_NEAREST) is still used to preserve discrete class labels and is therefore not applicable to the bounding-box detection task considered in this study [13].

Table 1 summarizes the data augmentation setups evaluated in this study. The evaluated setups and their associated parameter values were systematically selected to isolate the contribution of different augmentation options while maintaining a manageable experimental design. The No Augmentation Setup (NAS) serves as the baseline without augmentation. The Geometric Augmentation Setup (GAS) introduces the effect of commonly used geometric transformations alone, whereas the Geometric + Mosaic Augmentation Setup (GMAS) adds the additional contribution of Mosaic augmentation beyond geometric transformations. Finally, the Default Augmentation Setup (DAS) allows the comparison against the default data augmentation implemented in YOLO26.

*2.2. YOLO26 Model Size Variants*

The YOLO26 model family comprises five size variants: nano (YOLO26n), small (YOLO26s), medium (YOLO26m), large (YOLO26l), and extra-large (YOLO26x) [6]. These variants share the same underlying architecture and training framework while differing in model capacity and computational complexity. All five variants support the same computer vision tasks, including object detection, instance segmentation, image classification, pose estimation, and oriented object detection [6]. The availability of multiple model sizes enables evaluation across different computational budgets while maintaining a consistent architectural design. In this study, all five YOLO26 model size variants were trained and evaluated using identical experimental settings to systematically assess their suitability for adenovirus detection [28].

*2.3. TEM virus dataset*

The TEM virus dataset contains a total of 1245 images of 22 virus classes captured with two different electron microscopes: an LEO (Zeiss, Oberkochen, Germany) with a Morada (Olympus) camera and a Tecnai 10 (FEI, Hillsboro, OR, USA) with a MegaView III (Olympus, Münster, Germany) camera [7]. The virus classes in the dataset are strongly unbalanced both regarding the number of images (from 9 to 129) and in the number of virus particles (from 38 to 1934). The sizes of all raw images are either 1376 x 1032 or 2048 x 2048 pixels (depending on with which electron microscope they were captured) but the pixel sizes vary from 0.26 to 5.57 nm, i.e., they were acquired at different magnifications [7]. Each virus particle was annotated only with its approximate center, i.e., a single point for isolated spherical particles or a centerline in case of clustered viruses (beyond visual recognition of individual particles) or elongated viruses. The annotations were in the form of coordinate points stored in a separate text file for each image [7]. Although the published TEM virus dataset contains 22 virus classes, only the adenovirus class was used in this study. The raw adenovirus dataset comprised of train set (40 images containing 160 particles), test set (14 images containing 86 particles) and validation set (13 images containing 44 particles) [3, 7].

*2.4. Computational Environment*

All training experiments were conducted on a SLURM-managed High-Performance Computing (HPC) cluster belonging to the University of Limerick and equipped with NVIDIA A100 SXM4-40GB GPUs. GPU resource utilization and system performance were monitored using NVIDIA's nvidia-smi utility, which provides real-time statistics on GPU memory usage, processing load, temperature, power consumption, CUDA environment, and active computational processes throughout the training and inference stages. No additional environment modules or containerized runtime were used during the experiments. The training and evaluation script was developed in Python and publicly available via Zenodo repository [28].

*2.5. Annotations*

Annotation bounding boxes were generated from the adenovirus particle centre coordinates provided in the TEM adenovirus dataset. Since the original annotations consisted only of particle centre coordinates [7], a custom Python script was developed to automatically generate YOLO-compatible annotation files [29]. For each annotated particle, a fixed-size bounding box (60 × 60 pixels) was centred on the recorded particle position. Bounding boxes extending beyond the image boundaries were clipped to the valid image region before conversion to the normalized YOLO annotation format according to the standard YOLO representation shown in Eq. (4), [35].

$$x_n = \frac{\hat{x}_{\min} + \hat{x}_{\max}}{2W},\ y_n = \frac{\hat{y}_{\min} + \hat{y}_{\max}}{2H}$$
$$w_n = \frac{\hat{x}_{\max} - \hat{x}_{\min}}{W},\ h_n = \frac{\hat{y}_{\max} - \hat{y}_{\min}}{H} \quad (4)$$

Thus, each annotation was stored as $(c, x_n, y_n, w_n, h_n)$ where $c = 0$ denotes the adenovirus class.

*2.6. Evaluation metrics*

The adenovirus detection performance was quantitatively evaluated using *Precision*, *Recall*, *mAP@50*, and *mAP@50–95*. *Precision* and *Recall* were computed as shown in Eq. (5) and Eq. (6), respectively [4].

$$\text{Precision} = \frac{TP}{TP + FP} \tag{5}$$

$$\text{Recall} = \frac{TP}{TP + FN} \tag{6}$$

As can be understood, the *Precision* emphasizes reducing false positives, whereas *Recall* emphasizes reducing false negatives [30]. A predicted bounding box $B_p$ was matched to a ground-truth box $B_g$ according to their Intersection over Union as shown in Eq. (7) [36], [37].

$$IoU(B_p, B_g) = \frac{|B_p \cap B_g|}{|B_p \cup B_g|} \tag{7}$$

The *mAP@50* metric denotes the mean Average Precision computed at an Intersection over Union (IoU) threshold of 0.50, while *mAP@50–95* denotes the mean Average Precision averaged over ten IoU thresholds ranging from 0.50 to 0.95 in increments of 0.05 [31].

### *2.7. YOLO26 Modifications*

The modified YOLO26 detector was developed as a baseline or general-purpose object detector - using the default *TaskAlignedAssigner*, automatic optimizer selection, and P3–P5 detection feature maps - is not specifically optimized for every application. To better accommodate small adenovirus particles found in TEM images, four modifications were introduced: (i) expansion of the STAL region during candidate selection, (ii) increasing the number of positive samples assigned by the *TaskAlignedAssigner* (TAL), (iii) replacing the automatically selected optimizer (i.e., AdamW) with MuSGD, and (iv) leveraging P2 feature maps. The remaining network architecture and training procedures were unchanged.

#### 2.7.1. *Small Target Label Assignment Expansion*

During candidate selection, the width and height of each valid ground-truth box were examined independently. Any dimension smaller than 64 pixels was expanded to 64 pixels, while dimensions already equal to or larger than 64 pixels were retained. Thus, the original 60 × 60-pixel adenovirus boxes were represented by 64 × 64-pixel candidate regions. This expansion was used only to determine whether prediction-point centres fell inside the ground-truth candidate region; the original boxes remained available for the subsequent assignment and bounding-box regression operations.

#### 2.7.2. *Positive Sample Assignment*

The YOLO26 detector employs the TAL for one-to-many positive-sample assignment during training, following the TAL strategy proposed in [38]. Let $k_r$ denote the number of positive candidates requested by an assignment operation. To increase the number of candidate predictions considered during one-to-many positive-sample assignment, this parameter was modified according to Eq. (8)

$$k = \begin{cases} 15, & k_r > 1 \\ k_r, & k_r \leq 1 \end{cases} \tag{8}$$

Where a secondary assignment parameter $k_2$ was present, it was constrained by the proposed rule in Eq. (9) to maintain consistency with the modified assignment strategy.

$$k_2' = min(k_2, 15) \tag{9}$$

This increased the maximum number of candidate predictions considered during the one-to-many positive-sample assignment.

2.7.3. *Optimization*

The original YOLO26 training framework automatically selects the default optimizer. In the baseline model, the automatically selected optimizer was *AdamW*. In this study, the optimizer was explicitly set to *MuSGD*, while all remaining training parameters were kept unchanged.

2.7.4. *Feature Representation*

Here, the standard P3–P5 detection hierarchy was replaced by the P2–P5 to introduce an additional high-resolution P2 detection head for improved small-object representation [32],[33]. For each model size variant, the corresponding official P2 YAML architecture (yolo26n-p2.yaml–yolo26x-p2.yaml) was instantiated, after which compatible pretrained parameters from the corresponding standard YOLO26 model were transferred before fine-tuning on the adenovirus dataset.

## 3. Results

The experimental results are organized into four subsections: (i) quantitative detection, (ii) training and inference time, (iii) YOLO26n convergence analysis, and (iv) qualitative detection.

### *3.1. Quantitative Detection*

#### *3.1.1 Baseline YOLO26*

Table 2 presents the adenovirus detection performance on the test set. Here, five YOLO26 size variants and four evaluated data augmentation setups were evaluated using *mAP@50–95*, *mAP@50*, *Precision*, and *Recall* metrics. As shown in Table 2, the baseline YOLO26n trained with GAS achieved the highest overall detection accuracy, with an *mAP@50–95* of 0.44, an *mAP@50* of 0.69, and a *Precision* of 0.86.

Table 2: Test-set detection performance of baseline YOLO26 size variants (SV) under different data augmentation sets.

| ***SV*** | ***mAP@50-95*** | | | | ***mAP@50*** | | | | ***Precision*** | | | | ***Recall*** | | | |
|---|---|---|---|---|---|---|---|---|---|---|---|---|---|---|---|---|
| | ***DAS*** | ***NAS*** | ***GAS*** | ***GMAS*** | ***DAS*** | ***NAS*** | ***GAS*** | ***GMAS*** | ***DAS*** | ***NAS*** | ***GAS*** | ***GMAS*** | ***DAS*** | ***NAS*** | ***GAS*** | ***GMAS*** |
| *26n* | 0.40 | 0.39 | 0.44 | 0.31 | 0.59 | 0.54 | 0.69 | 0.68 | 0.68 | 0.81 | 0.86 | 0.75 | 0.65 | 0.52 | 0.61 | 0.56 |
| *26s* | 0.34 | 0.39 | 0.28 | 0.26 | 0.49 | 0.57 | 0.64 | 0.61 | 0.61 | 0.81 | 0.72 | 0.65 | 0.49 | 0.56 | 0.52 | 0.56 |
| *26m* | 0.29 | 0.35 | 0.20 | 0.28 | 0.44 | 0.52 | 0.44 | 0.56 | 0.61 | 0.78 | 0.72 | 0.82 | 0.52 | 0.52 | 0.45 | 0.48 |
| *26l* | 0.33 | 0.36 | 0.20 | 0.26 | 0.51 | 0.52 | 0.48 | 0.58 | 0.65 | 0.79 | 0.66 | 0.68 | 0.52 | 0.46 | 0.46 | 0.48 |
| *26x* | 0.38 | 0.39 | 0.22 | 0.29 | 0.60 | 0.55 | 0.54 | 0.64 | 0.71 | 0.81 | 0.76 | 0.74 | 0.64 | 0.52 | 0.46 | 0.59 |

However, the highest *Recall* of 0.65 was obtained using the DAS. The effect of augmentation varied across the model size variants. GAS was most effective for YOLO26n, whereas GMAS produced the highest *mAP@50* values for YOLO26m, YOLO26l, and YOLO26x. NAS yielded the highest *mAP@50–95* for YOLO26s, YOLO26m, and YOLO26l, while YOLO26x obtained the same highest *mAP@50–95* value of 0.39 with both NAS and DAS. Overall, increasing the model size did not result in a corresponding improvement in detection performance, as the baseline *nano* variant achieved the highest values, in general, across the experiments.

#### *3.1.2 Modified YOLO26*

As shown in Table 3, the modified YOLO26x trained with GAS achieved the highest mAP@50–95 of 0.49 and the highest Recall of 0.80, while attaining an mAP@50 of 0.80 and a Precision of 0.81. The highest Precision of 0.93 was achieved by YOLO26m trained with GMAS.

Table 3: Test-set detection performance of modified YOLO26 size variants (SV) under different data augmentation sets.

| *SV* | ***mAP@50-95*** | | | | ***mAP@50*** | | | | ***Precision*** | | | | ***Recall*** | | | |
|---|---|---|---|---|---|---|---|---|---|---|---|---|---|---|---|---|
| | ***DAS*** | ***NAS*** | ***GAS*** | ***GMAS*** | ***DAS*** | ***NAS*** | ***GAS*** | ***GMAS*** | ***DAS*** | ***NAS*** | ***GAS*** | ***GMAS*** | ***DAS*** | ***NAS*** | ***GAS*** | ***GMAS*** |
| *26n* | *0.39* | 0.37 | 0.47 | 0.45 | *0.59* | 0.53 | 0.75 | 0.77 | *0.56* | 0.81 | 0.88 | 0.90 | *0.64* | 0.52 | 0.69 | 0.61 |
| *26s* | *0.29* | 0.38 | 0.45 | 0.49 | *0.47* | 0.53 | 0.71 | 0.80 | *0.69* | 0.66 | 0.77 | 0.87 | *0.48* | 0.60 | 0.68 | 0.76 |
| *26m* | *0.33* | 0.38 | 0.44 | 0.46 | *0.53* | 0.53 | 0.72 | 0.80 | *0.64* | 0.83 | 0.82 | 0.93 | *0.50* | 0.53 | 0.69 | 0.75 |
| *26l* | *0.33* | 0.38 | 0.33 | 0.45 | *0.47* | 0.53 | 0.56 | 0.75 | *0.79* | 0.90 | 0.80 | 0.85 | *0.50* | 0.48 | 0.51 | 0.64 |
| *26x* | *0.37* | 0.37 | 0.49 | 0.37 | *0.53* | 0.51 | 0.80 | 0.70 | *0.63* | 0.91 | 0.81 | 0.81 | *0.65* | 0.49 | 0.80 | 0.64 |

The effect of data augmentation varied across the model size variants. GMAS produced the highest mAP@50 values for four of the five model size variants and the highest mAP@50–95 values for YOLO26s, YOLO26m, and YOLO26l. GAS produced the best individual configuration, with modified YOLO26x + GAS achieving the highest mAP@50–95 and Recall. Unlike the baseline detector, the modified detector no longer exhibited the nano variant as the strongest performer, with the modified YOLO26x trained using GAS achieving the highest mAP@50–95 and Recall among the evaluated configurations.

#### *3.1.3 Baseline versus Modified YOLO26*

From Table 4, the improvements were not uniformly distributed across all augmentation strategies. Instead, they showed that the modifications consistently benefited the geometric augmentation setups (GAS and GMAS) much more than the default (DAS) and no-augmentation (NAS) setups. GAS exhibited the largest and most consistent improvements across all YOLO26 size variants. GMAS also showed substantial gains. In contrast, DAS and NAS exhibited only marginal improvements and, in several cases, slight decreases, suggesting that the proposed modifications provided limited additional benefit under these augmentation settings. This suggests that the modified detector was particularly effective at exploiting the greater geometric diversity introduced by GAS and GMAS.

Table 4: Improvement of the modified YOLO26 over the baseline YOLO26 under different data augmentation sets

| *SV* | ***mAP@50-95*** | | | | ***mAP@50*** | | | | ***Precision*** | | | | ***Recall*** | | | |
|---|---|---|---|---|---|---|---|---|---|---|---|---|---|---|---|---|
| | ***DAS*** | ***NAS*** | ***GAS*** | ***GMAS*** | ***DAS*** | ***NAS*** | ***GAS*** | ***GMAS*** | ***DAS*** | ***NAS*** | ***GAS*** | ***GMAS*** | ***DAS*** | ***NAS*** | ***GAS*** | ***GMAS*** |
| *26n* | -0.01 | -0.02 | +0.03 | +0.14 | +0.00 | -0.01 | +0.06 | +0.09 | -0.12 | +0.00 | +0.02 | +0.15 | -0.01 | +0.00 | +0.08 | +0.05 |
| *26s* | -0.05 | -0.01 | +0.17 | +0.23 | -0.02 | -0.04 | +0.07 | +0.19 | +0.08 | -0.15 | +0.05 | +0.22 | -0.01 | +0.04 | +0.16 | +0.20 |
| *26m* | +0.04 | +0.03 | +0.24 | +0.18 | +0.09 | +0.01 | +0.28 | +0.24 | +0.03 | +0.05 | +0.10 | +0.11 | -0.02 | +0.01 | +0.24 | +0.27 |
| *26l* | +0.00 | +0.02 | +0.13 | +0.19 | -0.04 | +0.01 | +0.08 | +0.17 | +0.14 | +0.11 | +0.14 | +0.17 | -0.02 | +0.02 | +0.05 | +0.16 |
| *26x* | -0.01 | -0.02 | +0.27 | +0.08 | -0.07 | -0.04 | +0.26 | +0.06 | -0.08 | +0.10 | +0.05 | +0.07 | +0.01 | -0.03 | +0.34 | +0.05 |

### *3.2. Training and Inference Time*

#### *3.2.1 Baseline YOLO26*

Table 5 compares the training and inference times of the five YOLO26 size variants under the four evaluated data augmentation setups. As expected, both training and inference times generally increased with model size, with YOLO26n requiring the shortest execution times and YOLO26x the longest. The effect of the data augmentation setups on computational time was less consistent across the evaluated models. GAS and GMAS frequently reduced the training time compared with the DAS and NAS. For example, YOLO26x required 29.31 minutes for training with the DAS, whereas GMAS reduced the training time to 20.59 minutes. In contrast, inference times exhibited only minor variations across the augmentation setups for a given model variant. Overall, model size had a substantially greater influence on inference time than the choice of data augmentation setup.

Table 5: Baseline YOLO26: Training and inference time comparison under different data augmentation sets

| *SV* | *Training time (minutes)* | | | | *Inference time (ms/image)* | | | |
|---|---|---|---|---|---|---|---|---|
| | **DAS** | **NAS** | **GAS** | **GMAS** | **DAS** | **NAS** | **GAS** | **GMAS** |
| *26n* | 9.88 | 10.71 | 9.61 | 6.79 | 6.57 | 7.76 | 8.08 | 8.47 |
| *26s* | 14.61 | 9.31 | 8.6 | 8.77 | 7.77 | 8.36 | 8.63 | 9.22 |
| *26m* | 16.94 | 16.21 | 9.91 | 10.85 | 11.45 | 10.69 | 10.65 | 11.08 |
| *26l* | 20.39 | 17.17 | 11.18 | 20.22 | 13.98 | 13.70 | 13.61 | 12.98 |
| *26x* | 29.31 | 29.27 | 24.14 | 20.59 | 22.21 | 21.99 | 22.45 | 21.92 |

*3.2.2 Modified YOLO26*

Table 6 compares the training and inference times of the five modified YOLO26 size variants under the four evaluated data augmentation setups. Similar to the baseline detector, both training and inference times generally increased with model size, with YOLO26n requiring the shortest execution times and YOLO26x the longest. However, unlike the baseline detector, no augmentation setup consistently reduced the training time across all model size variants. Instead, the effect of data augmentation on training time varied considerably depending on the model variant. Inference times also exhibited relatively small variations across the augmentation setups for a given model size, indicating that model size remained the dominant factor influencing inference speed.

Table 6: Modified YOLO26: Training and inference time comparison under different data augmentation sets

| *SV* | *Training time (minutes)* | | | | *Inference time (ms/image)* | | | |
|---|---|---|---|---|---|---|---|---|
| | **DAS** | **NAS** | **GAS** | **GMAS** | **DAS** | **NAS** | **GAS** | **GMAS** |
| *26n* | 12.68 | 16.61 | 14.41 | 13.57 | 8.78 | 10.79 | 12.31 | 10.30 |
| *26s* | 11.72 | 16.03 | 15.58 | 16.05 | 13.89 | 11.74 | 10.74 | 11.82 |
| *26m* | 16.71 | 21.61 | 21.69 | 21.54 | 14.25 | 15.92 | 16.20 | 18.28 |
| *26l* | 18.58 | 27.06 | 19.86 | 27.38 | 24.01 | 18.70 | 18.22 | 17.30 |
| *26x* | 31.82 | 34.54 | 37.37 | 34.48 | 37.87 | 32.68 | 32.71 | 32.64 |

*3.2.3 Runtime Comparison Baseline versus Modified YOLO26*

Table 7 summarizes the training and inference runtime differences between the modified and baseline YOLO26 detectors under the evaluated data augmentation setups. A positive value means the modified YOLO26 is slower than the baseline, while a negative value means it is faster. Overall, the detector modifications generally increased both training and inference times, although the magnitude of the increase varied across model size variants and data augmentation setups.

Table 7: Runtime differences between the modified and baseline YOLO26 detectors under the evaluated data augmentation setups

| *SV* | *Training time (minutes)* | | | | *Inference time (ms/image)* | | | |
|---|---|---|---|---|---|---|---|---|
| | **DAS** | **NAS** | **GAS** | **GMAS** | **DAS** | **NAS** | **GAS** | **GMAS** |
| *26n* | +2.80 | +5.90 | +4.80 | +6.78 | +2.21 | +3.03 | +4.23 | +1.83 |
| *26s* | -2.89 | +6.72 | +6.98 | +7.28 | +6.12 | +3.38 | +2.11 | +2.60 |
| *26m* | -0.23 | +5.40 | +11.78 | +10.69 | +2.80 | +5.23 | +5.55 | +7.20 |
| *26l* | -1.81 | +9.89 | +8.68 | +7.16 | +10.03 | +5.00 | +4.61 | +4.32 |
| *26x* | +2.51 | +5.27 | +13.23 | +13.89 | +15.66 | +10.69 | +10.26 | +10.72 |

### 3.3. *Baseline and Modified YOLO26n Convergence Analysis*

The baseline YOLO26n was selected for convergence analysis because it achieved the highest overall adenovirus detection performance among the evaluated baseline YOLO26 model size variants. For comparison purposes, the modified YOLO26n was also selected for convergence analysis. Figure 1 illustrates the training convergence of the baseline and modified YOLO26n under the four evaluated data augmentation setups. As shown in Figure 1, all four data augmentation setups exhibited successful convergence during training for both the baseline and modified YOLO26n, with the training and classification losses decreasing rapidly during the early epochs before gradually approaching relatively stable values. Similarly, Precision, Recall, mAP@50, and mAP@50–95 increased sharply during the initial training epochs before gradually reaching plateaus. For the baseline detector, GAS exhibited comparatively smoother convergence curves throughout the training process, whereas GMAS showed larger fluctuations in the validation losses and detection metrics, particularly for mAP@50–95. The modified YOLO26n generally converged to higher final detection metrics under GAS and GMAS, whereas its performance under DAS and NAS was comparable to or slightly lower than that of the corresponding baseline detector. Overall, all training runs converged under the evaluated data augmentation setups, although differences in convergence behaviour and final detection performance were observed between the baseline and modified detectors.

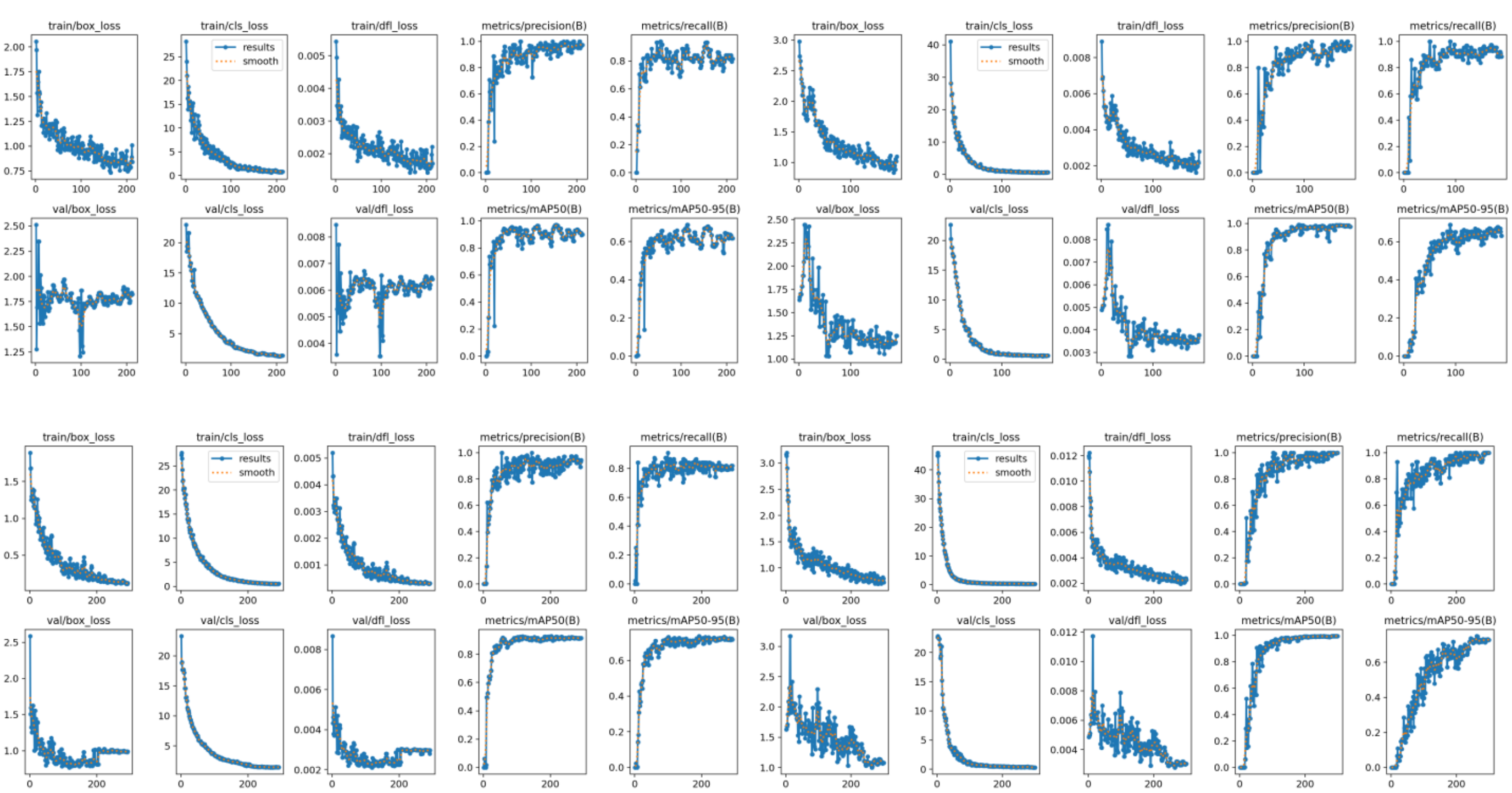

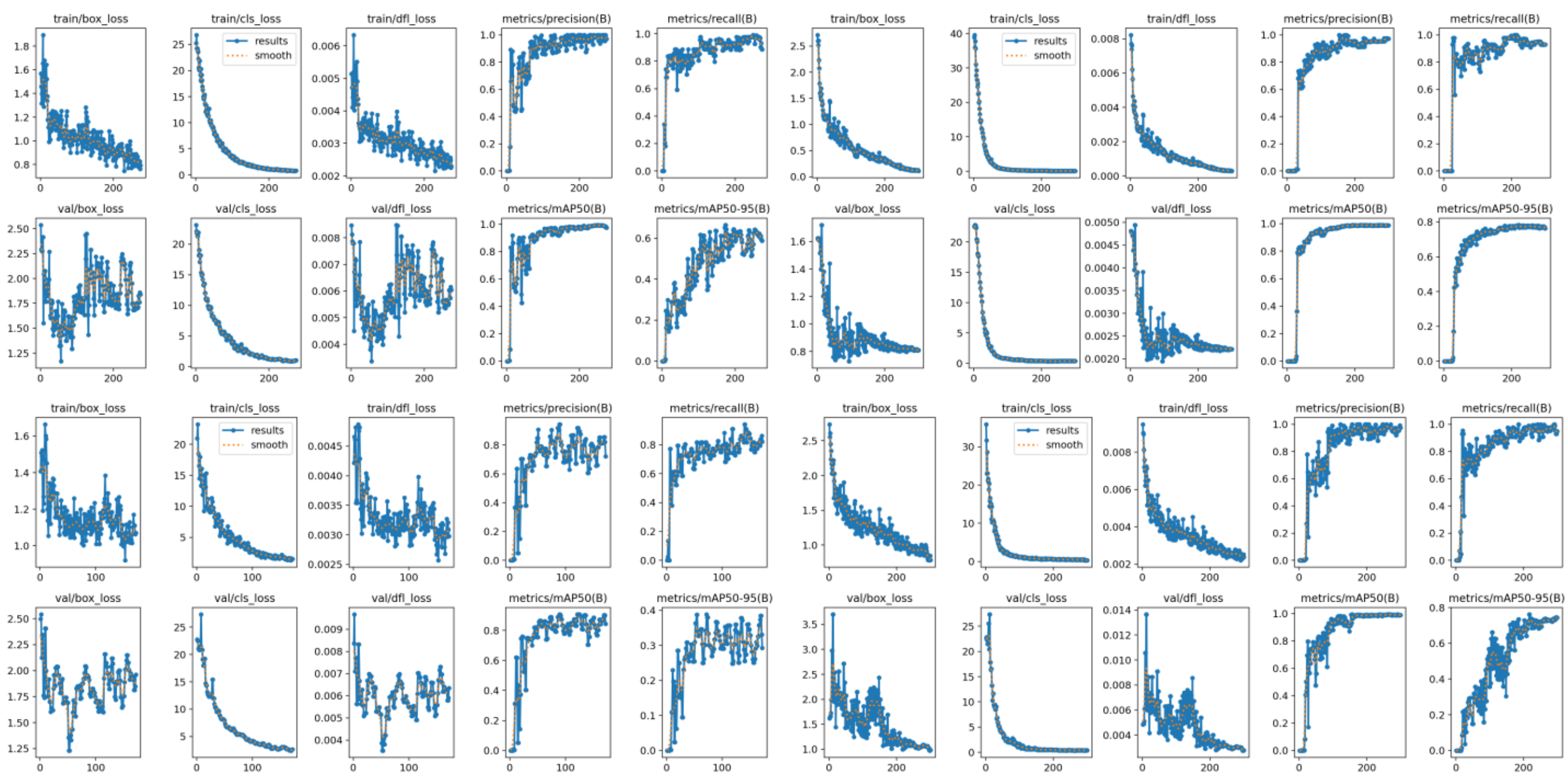

Figure 1: Comparison of training convergence curves under different data augmentation setups. Baseline YOLO26n (left column), modified YOLO26n (right column), DAS (row 1), GAS (row 2), NAS (row 3) and GMAS (row 4).

*3.4. Qualitative Detection*

Figure 2 presents representative qualitative adenovirus detection results produced by the baseline and modified YOLO26n, under four data augmentation setups using the selected four TEM test images.

Under the DAS setup, the baseline YOLO26 detector and the modified YOLO26 detector produced highly similar qualitative detection results. Both detectors correctly detected all adenoviruses in the first three images without generating false positives. In the fourth image, both detectors missed the same upper-left adenovirus while correctly detecting the remaining two adenoviruses, resulting in one false negative. Consequently, these representative examples do not reveal a clear qualitative difference between the baseline YOLO26 detector and the modified YOLO26 detector under the DAS setup, although the quantitative evaluation demonstrated performance differences across the complete test dataset.

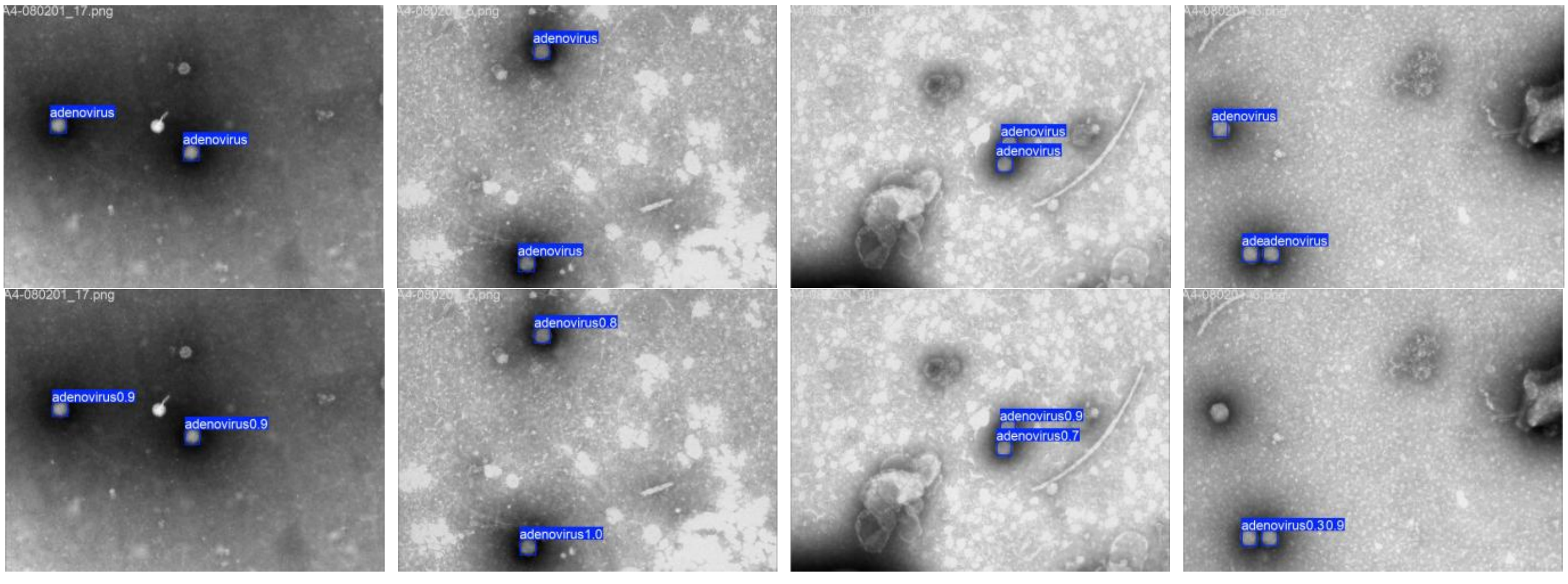

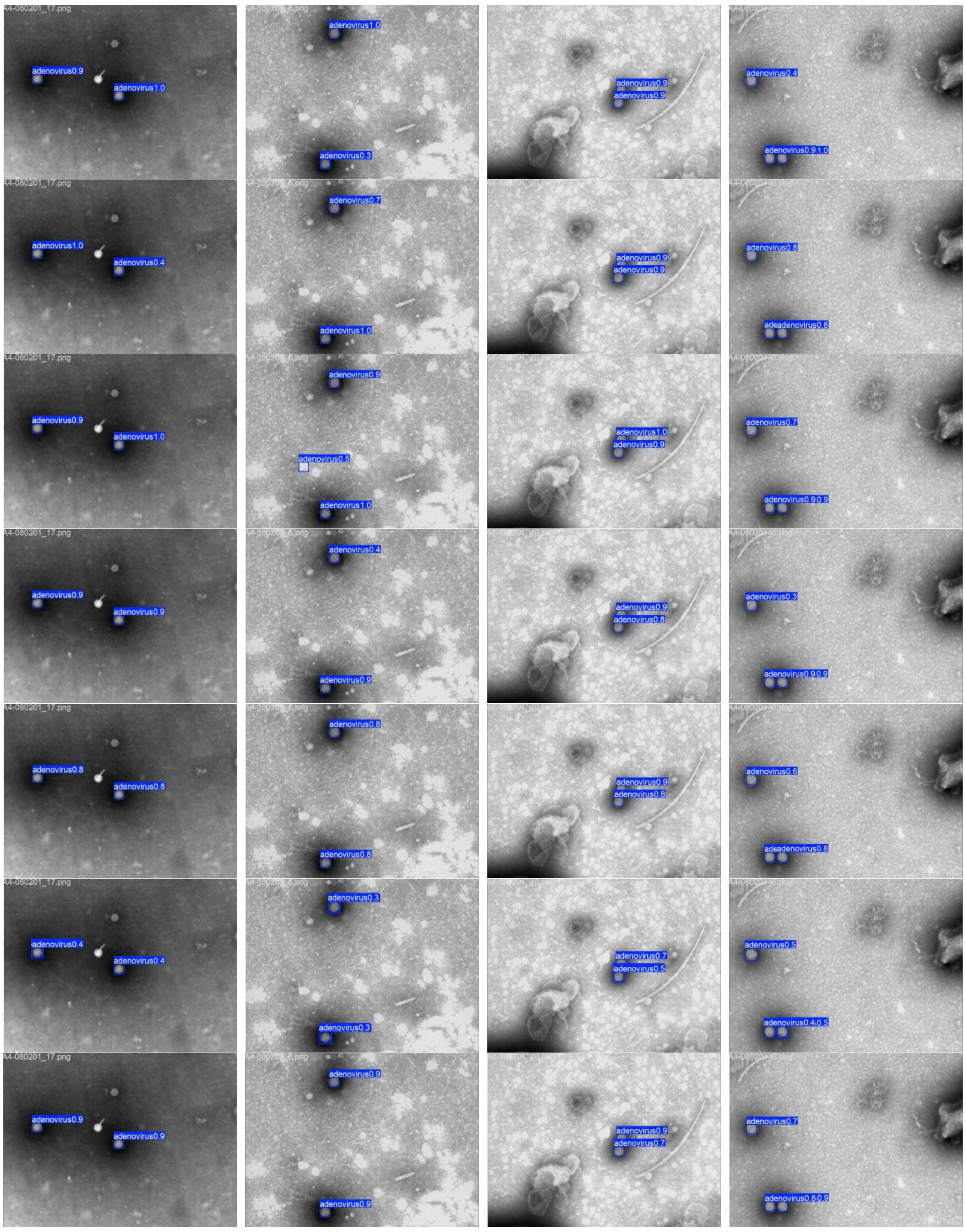

Figure 2: Comparison of adenovirus detections produced by the baseline and modified YOLO26n detectors. Ground-truth (first row). DAS [baseline YOLO26n (**second**) and modified YOLO26n (**third**) rows], NAS [baseline YOLO26n (**fourth**) and modified YOLO26n (**fifth**) rows], GAS [baseline YOLO26n (**sixth**) and modified YOLO26n (**seventh**) rows], and GMAS [baseline YOLO26n (**eighth**) and modified YOLO26n (**ninth**) rows].

Under the NAS setup, the baseline YOLO26 detector and the modified YOLO26 detector produced highly similar qualitative detection results across the representative images. Both detectors correctly detected all adenoviruses in Images 1 and 3 without generating false positives and exhibited the same missed detection in Image 4. The primary qualitative difference was observed in Image 2, where the modified YOLO26 detector generated one additional false positive while still correctly detecting both ground-truth adenoviruses. Overall, these representative examples indicate comparable qualitative detection performance between the baseline YOLO26 detector and the modified YOLO26 detector under the NAS setup, with only a minor increase in false-positive detections observed for the modified detector in one example. Under the GAS setup, the baseline YOLO26 detector and the modified YOLO26 detector produced virtually identical qualitative detection results across the representative images. Both detectors correctly detected all adenoviruses in the first three images without generating false positives and exhibited the same missed detection in the fourth image. Consequently, these representative examples do not reveal a qualitative difference between the baseline YOLO26 detector and the modified YOLO26 detector under the GAS setup, although the quantitative evaluation demonstrated that GAS provided the largest performance improvements across the complete test dataset. Under the GMAS setup, the baseline YOLO26 detector and the modified YOLO26 detector produced nearly identical qualitative detection results across the representative images. Both detectors correctly detected all adenoviruses in the first three images without generating false positives and exhibited the same missed detection in the fourth image. Consequently, these representative examples do not reveal a clear qualitative difference between the baseline YOLO26 detector and the modified YOLO26 detector under the GMAS setup, although the quantitative evaluation demonstrated that the modified detector achieved superior overall performance under the GMAS augmentation strategy across the complete test dataset. Overall, the modified YOLO26 detector generally produced higher confidence scores for correctly detected adenoviruses than the baseline YOLO26 detector across the evaluated augmentation setups. Note that these observations are specific to these four representative TEM images shown in Figure 2 and are intended solely to support the qualitative assessment rather than provide a quantitative comparison across the complete test dataset.

## 4. Discussion

Our discussion focuses on the effects of the evaluated data augmentation setups, the performance of the baseline and modified YOLO26 detectors for adenovirus detection, the practical implications of the findings, and the study limitations and future research directions.

### 4.1. Effect of Data Augmentation Setups

The experimental results demonstrated that the effectiveness of data augmentation depended on the augmentation setup employed. For the modified detector, GMAS produced the most consistently high performance across model size variants, whereas GAS produced the single best-performing configuration, modified YOLO26x + GAS, and generally yielded the largest improvements relative to the baseline detector. Furthermore, the detector modifications benefited the geometric augmentation setups (GAS and GMAS) considerably more than the default augmentation (DAS) and no-augmentation (NAS) setups. These findings suggest that the combined modifications - P2 feature representation, expanded STAL, increased positive-sample assignment, and MuSGD - were particularly effective when used with GAS and GMAS. Because the modifications were evaluated jointly, the contribution of each individual modification cannot be isolated from the present experiments. Although Mosaic increased training-image diversity, GMAS did not consistently outperform its corresponding non-Mosaic geometric setup, GAS. Similarly, DAS did not consistently outperform NAS; however, these two setups differ in several default augmentation operations and therefore do not isolate the effect of Mosaic alone.

This suggests that combining multiple TEM images into a single training image does not necessarily provide additional discriminative information for adenovirus detection and may introduce image compositions that differ from the natural appearance of adenovirus particles. The qualitative evaluation further showed that the baseline and modified YOLO26 detectors produced comparable true-positive, false-positive, and false-negative detection patterns across the representative examples or images. However, the modified YOLO26 detector generally assigned higher confidence scores to correctly detected adenoviruses than the baseline YOLO26 detector, indicating greater confidence in its predictions while maintaining comparable qualitative detection performance.

### *4.2. Performance of the Baseline and Modified YOLO26 Detectors*

The experimental results demonstrated that the proposed detector modifications generally improved the performance of YOLO26 for adenovirus detection. However, the magnitude of the improvements depended on the selected data augmentation setup. The largest improvements were achieved under the GAS and GMAS setups, whereas the DAS and NAS setups exhibited only marginal improvements and, in some cases, slight decreases. These findings suggest that the proposed detector modifications are particularly effective when combined with augmentation setups that introduce greater geometric variability. Among the evaluated baseline YOLO26 detectors, YOLO26n achieved the highest overall detection performance. However, after applying the proposed detector modifications, YOLO26x under the GAS and GMAS augmentation setups achieved the highest overall detection performance. This shift from the lightweight YOLO26n baseline detector to the larger YOLO26x modified detector suggests that the proposed detector modifications enabled larger model size variants to better exploit the additional geometric information introduced by the GAS and GMAS augmentation setups. These findings indicate that optimal adenovirus detection depends on the combined effects of detector design, model size, and data augmentation setup.

### *4.3. Practical Implications*

The experimental findings indicate that improving adenovirus detection in TEM images requires an appropriate combination of detector design and data augmentation setup. Although the proposed YOLO26 modifications generally increased both training and inference times, they also improved detection performance, particularly under the GAS and GMAS setups. Consequently, combining the modified YOLO26 detector with geometric data augmentation set provided a practical approach for improving adenovirus detection with a measurable computational trade-off. Furthermore, the qualitative evaluation demonstrated that the modified YOLO26 detector generally produced higher confidence scores for correctly detected adenoviruses than the baseline YOLO26 detector, despite both detectors exhibiting comparable qualitative detection patterns in the representative examples.

### *4.4. Study Limitations and Future Work*

This study was conducted using the selected adenovirus TEM dataset. Therefore, the observed performance differences between the baseline and modified YOLO26 detectors, as well as the effectiveness of the evaluated data augmentation setups are only linked to this adenovirus dataset. Future research should validate these findings using additional virus datasets, investigate the effectiveness of the proposed detector modifications for other one-stage and transformer-based object detectors, and evaluate whether similar performance trends are observed under similar or different imaging conditions.

## 5. Conclusions

This study systematically benchmarked four data augmentation setups across five baseline and modified YOLO26 model size variants to investigate their effects on adenovirus detection in TEM images. The experimental results demonstrated that the proposed de-tector modifications generally improved adenovirus detection performance, with the largest improvements achieved under the geometric augmentation setups (GAS and GMAS). The findings further demonstrated that the proposed detector modifications changed the relationship between model size and detection performance. Whereas the baseline YOLO26n detector achieved the highest overall performance among the baseline models, the proposed modifications shifted the best-performing configurations toward the larger model variants, with the modified YOLO26x trained using GAS emerging as a particularly strong-performing configuration. These results indicate that optimal adenovirus detection depends on the combined effects of detector design, model size, and data augmentation rather than on any single factor alone. Overall, this benchmarking study provides practical guidance for selecting effective combinations toward optimal adenovirus detection using YOLO26 and establishes a foundation for future evaluation on larger and more diverse TEM datasets.

**Supplementary Materials:** All developed Python source codes related to this work can be publicly accessed via the Zenodo repository [28-29].

**Author Contributions:** Conceptualization, O.R.; methodology, O.R.; software, O.R.; validation, O.R.; formal analysis, O.R.; investigation, O.R.; resources, O.R.; data curation, O.R.; writing—original draft preparation, O.R.; writing—review and editing, O.R.; visualization, O.R.; supervision, O.R.; project administration, O.R.; funding acquisition, O.R. The author has read and agreed to the published version of the manuscript.

**Funding:** This research received no external funding.

**Data Availability Statement:** Data supporting this work are publicly available [7].

**Acknowledgments:** The author would like to thank the University of Eastern Finland and FinVector for encouragement.

**Conflicts of Interest:** The authors declare no conflicts of interest.